# The Digital Synaptic Neural Substrate:
# Size and Quality Matters

Azlan Iqbal
College of Computer Science and Information Technology, Universiti Tenaga Nasional
Putrajaya Campus, Jalan IKRAM-UNITEN, 43000 Kajang, Selangor, Malaysia
azlan@uniten.edu.my

ABSTRACT

We investigate the 'Digital Synaptic Neural Substrate' (DSNS) computational creativity approach further with respect to the size and quality of images that can be used to seed the process. In previous work we demonstrated how combining photographs of people and sequences taken from chess games between weak players can be used to generate chess problems or puzzles of higher aesthetic quality, on average, compared to alternative approaches. In this work we show experimentally that using larger images as opposed to smaller ones improves the output quality even further. The same is also true for using clearer or less corrupted images. The reasons why these things influence the DSNS process is presently not well-understood and debatable but the findings are nevertheless immediately applicable for obtaining better results.

## 1    INTRODUCTION

In previous work we have explained and demonstrated in some detail the 'Digital Synaptic Neural Substrate' (DSNS) as a generic approach in computational creativity that can be used to generate a virtually unlimited number of creative objects, using chess problems or puzzles as the domain of investigation (Iqbal et al., 2016). We will therefore not repeat any material that has already been covered. For the benefit of readers, we can summarize that the DSNS approach involves automatically analyzing objects (e.g. a chess problem or photograph) for their attributes (e.g. number of pieces, piece value or material difference, image resolution, the number of colors used) that can be represented in numerical form. The 'creative difference' (i.e. 'deviation value') between two such objects is then calculated using a simple formula in order to pseudo-randomly generate the attributes for two *new* theoretical objects that produce the same or similar deviation value. These differing attribute values can then be used by an external object-generation system to produce the two new theoretical 'child' objects that have the potential of being similar in creative value to their 'parents'. As a very simple example, if we start with two boxes that have only height, width and length measurements in inches as their attributes, i.e. say, 5 x 6 x 7 and 7 x 8 x 9, the DSNS process may produce two new sets of attributes such as 4.5 x 2.25 x 8 and 6 x 3 x 1 and these new measurements can be used to create or build two new boxes that may be as attractive as their parents.

In the aforementioned previous work (Iqbal et al., 2016) it was demonstrated that photographs of people (as opposed to say, paintings and computer-generated art) used in combination with sequences taken from chess games between weak players (as opposed to say, chess problems by experienced composers) produced the highest quality output via the DSNS approach. The reason for this remains an open question. The output referred to are computer-generated chess problems, as the output type can be from either one of the source domains used (in this case, photographs or chess problems). There are other open questions remaining as well such as why the DSNS approach should even work at all and we will not be attempting to address them here. Instead, we will focus on extending the work by demonstrating experimentally how the quality of the output can be further improved by taking into account the properties of the images used in the process. Section 2 explains the experimental design, section 3 presents a discussion of the results and section 4 concludes with some directions for further work.

## 2    EXPERIMENTAL DESIGN

The basic idea in this research was to test if using larger images and those of higher quality in the DSNS process as described in (Iqbal et al., 2016) could improve the quality of the output. For the experiments we used the Chesthetica (v9.99) computer program from earlier work (Iqbal et. al, 2016) to evaluate the aesthetics of all the chess problems it created. This was the criterion used in determining a better or higher quality object output from the system. Even though the difference between two aesthetic scores may be small, that difference nevertheless ranks one chess problem above the other. Chesthetica uses an experimentally-validated aesthetics model that is able to rank chess problems aesthetically in a way that correlates positively and well with domain-competent human assessment (Iqbal et al., 2012). It is the most cost-effective and reliable method of evaluating beauty in chess move sequences as opposed to say, using human experts whom are not only expensive but also physically and mentally incapable of evaluating thousands of such sequences consistently.

The photographs of people used in the DSNS process were licensed and selected from the Bigstock (2016) image archive whereas the chess sequences (i.e. three-move mate endings) from tournament games between weak players (Elo rating < 1500) were obtained randomly from the ChessBase (2016) Big Database 2011. The photographs were not selected with anything particular in mind except that there were clearly one or more people in them. Needless to say, such people tended to be attractive. To compare means, a two-sample F-test for variances was first applied to the samples to determine whether a two-sample, two-tailed, T-test assuming equal or unequal variances (i.e. TTEV or TTUV) should be used (at the 5% significance level). The computer used to perform the experiments was an Acer Notebook with an Intel(R) Core(TM) i3 CPU M 370 @ 2.40GHz, running Windows 7 Home Premium 64-bit with 4 GB of RAM.

2.1 Larger Images versus Smaller Images

The first experiment compared chess problems (forced mates in 4 or 5 moves; collectively known as 'long mates') generated by Chesthetica via the DSNS approach using 300 photographs of people and 1,000 tournament game sequences between weak players. The composing method is described in detail in (Iqbal et al., 2016). These numbers have no particular significance but were considered both manageable and sufficient given the resources available to us. The photographs used were all resized down from the original very high resolution versions to a width of 1,920 pixels by a height that maintained the original aspect ratio. The height for all the images was over 1,000 pixels regardless. Figure 1 shows two examples of such images, smaller than actually used in the experiment. The output of chess problems generated using this setup was considered the first set or sample (A) where the average aesthetic score for those problems would be compared against the average aesthetic score of another sample (B), using a slightly different setup as explained below.

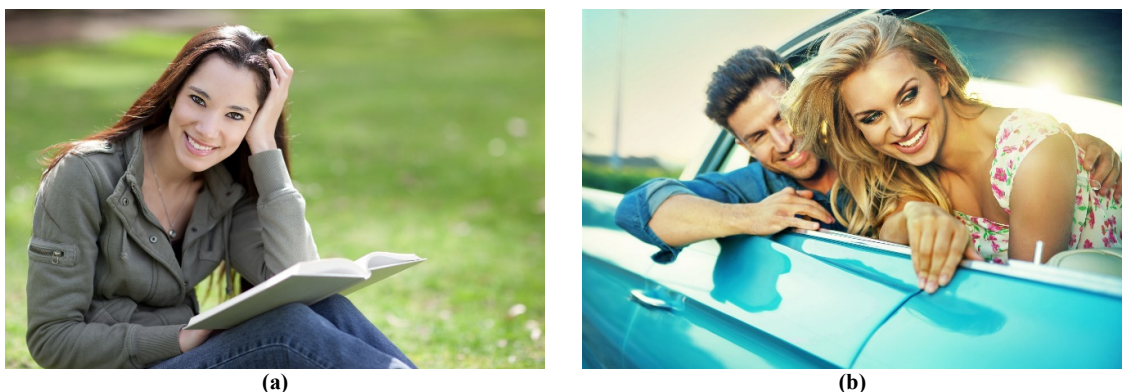

(a) (b)
**Figure 1**: Examples of the images used.

The second sample (B) of chess problems was generated via the DSNS approach using the same set of tournament games. However, while the same images were used here as well, they were resized from the originals to a height of 720 pixels by a width that maintained the original aspect ratio. The width for all the images was always below 1,920 pixels regardless. This guaranteed that the images in this case were smaller than the ones used in sample A. So the only distinguishing factor between the two samples of generated chess problems would be the size of the photographs used to seed the DSNS process. Chesthetica was set to compose problems for 30 cycles of 6 hours so that composing efficiency (i.e. compositions per hour for every cycle) could also be measured and the average composing efficiency over 30 cycles compared. Four typical composition conventions and being limited to only the original piece set (i.e. no promoted pieces) were specified in the program settings as a compromise between performance and quality (see Appendix A). There was no statistically significant difference in terms of average composing efficiency between the two samples, i.e. 1.062 compositions per hour (cph) for sample A and 1.072 cph for sample B. However, aesthetically, sample A scored, on average, 2.455 whereas sample B, on average, scored only 2.351 and this difference was indeed statistically significant; TTEV: t(374) = 2.093, P = 0.0367.

## 2.2 Clearer Images versus Corrupted Images

The second experiment was set up similar to the first (see section 2.1) except that instead of testing for the influence of image size, we tested for image clarity. The first sample in this experiment used the same 300 images but resized to a width of 800 pixels and a height that maintained the original aspect ratio. This was done because of certain software limitations related to later corrupting those same images. The quality of these images were therefore same as those shown in Figure 1. Chess problems were also generated using this setup. The second sample had the same images corrupted slightly using the Apple iPad 'Glitch' app. This program uses a variety of methods to introduce interference into a photo such as the random lines shown in Figure 2. The level of corruption (termed here as 'level 1') introduced was intended to keep the images still recognizable to humans.

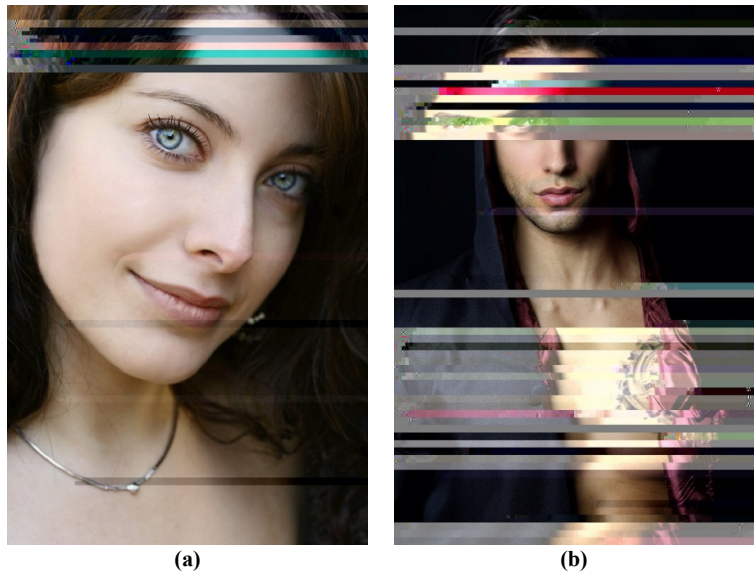

(a) (b)

**Figure 2**: Examples of the images (level 1 corruption).

Chess problems were generated using this setup as well and combined with the previous ones (generated with no corruption in the seed images). Together, this was the 'clearer image' sample (C). The reason the output using these was combined with the output using the totally uncorrupted photos was to provide a more realistic scenario given that in most real-world situations, having all photographs in a collection absolutely pristine and clear is highly unlikely. Most images, especially those taken directly using standard cameras and smartphones under non-perfect lighting conditions, are not as clear as the Bigstock photos. We had considered using a random 'grain' sort of interference instead of lines but that feature was not available using any software package we could find. The vast majority of image editing programs focus on enhancing photos, not corrupting them using random algorithms.

For the 'corrupted image' sample (D), the same uncorrupted photos at a width of 800 pixels were subjected to two higher levels of corruption (levels 2 and 3) using the 'Glitch' app. These were intended to make them barely recognizable and virtually unrecognizable. Figures 3 and 4 show examples. In Figure 3, a swirling-type of interference was introduced at random into each image whereas in Figure 4, a more intense blocky or mosaic random interference was introduced. The pictures may nevertheless become more recognizable to human eyes if viewed at a distance. Experimentally, there was no statistically significant difference in terms of average composing efficiency between the two samples C and D, i.e. 1.030 cph and 1.067, respectively. However, aesthetically, sample C scored, on average, 2.396 whereas sample D scored, on average, 2.330 and this difference was indeed also statistically significant; TTEV: $t(756) = 1.993$, $P = 0.0466$.

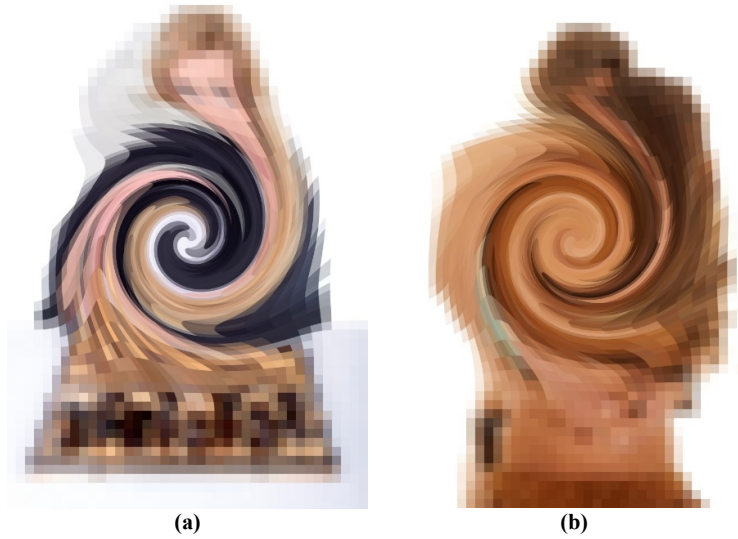

**Figure 3**: Examples of the images (level 2 corruption).

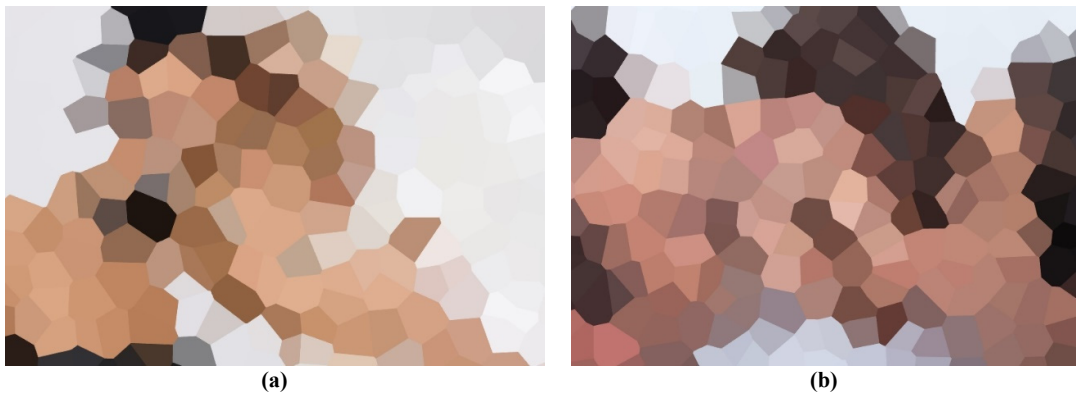

**Figure 4**: Examples of the images (level 3 corruption).

## 3   DISCUSSION

The experimental results suggest that using larger images and those of higher quality (i.e. less corrupted) actually improve the output of the DSNS process in terms of the aesthetic quality of the chess problems generated. This is a new or extended finding to our earlier work and highly relevant to the implementation of the DSNS approach in any domain of creative endeavor. Figure 5 shows one example each of an average-scoring chess problem generated using the larger images and smaller images as explained in section 2.1. The line or move sequence shown below the board is the main winning variation. Note that the aesthetic score for each of the problems shown in Figure 5 is the closest to the average score obtained in their sample during experimentation (refer section 2.1). This should give the reader some idea of the difference in quality that can be achieved by taking into account image size. Most chess players, at least according to the experimentally-validated aesthetics model incorporated into Chesthetica, would agree that the problem in (a) is of higher aesthetic quality than (b).

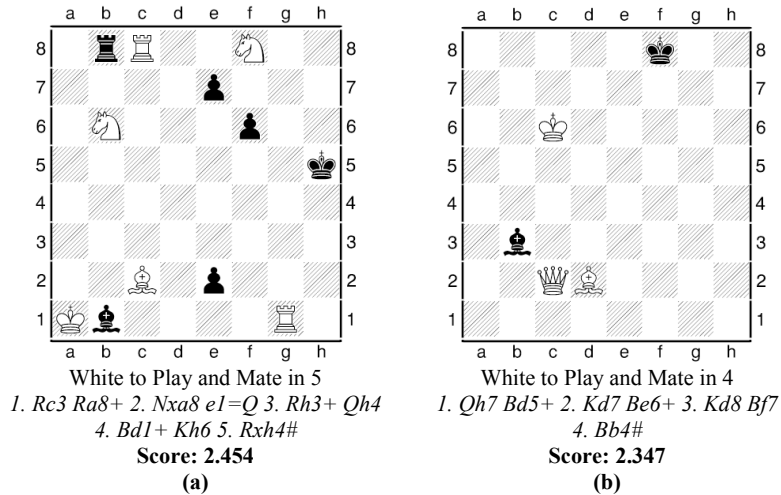

White to Play and Mate in 5
1. Rc3 Ra8+ 2. Nxa8 e1=Q 3. Rh3+ Qh4
4. Bd1+ Kh6 5. Rxh4#
**Score: 2.454**
**(a)**

White to Play and Mate in 4
1. Qh7 Bd5+ 2. Kd7 Be6+ 3. Kd8 Bf7
4. Bb4#
**Score: 2.347**
**(b)**

**Figure 5**: Example of an average-scoring composition generated using larger images (a) and smaller ones (b).

On the surface, it might appear to make sense that larger images, simply by virtue of having more pixels or information, possibly provide more 'working space' for the DSNS in the computational creativity process; and that this somehow facilitates improving the quality of the compositions generated. However, it is difficult to reconcile that idea with the finding that simply using *clearer* images (or less corrupted ones) actually improves the process as well. Humans might find corrupted and highly corrupted images (such as those in Figures 3 and 4) to be less 'inspiring' because their content cannot be easily recognized (if at all), but these images contain the same number of pixels as the clearer images they were compared against. So the DSNS process appears to be 'recognizing' the content of the images somehow, if an analogy to human inspiration is to be made. It may be that clearer images have a certain 'harmony' about them that affects how easy it is to randomly find new DSNS 'child' attributes that more closely resemble their 'parent' attributes. Regardless of the actual reason behind these phenomena, the experimental results are useful and immediately applicable in the implementation of the DSNS approach to generate creative objects.

In this research, we were able to confirm these results in the domain of chess problem composition thanks, in part, to the availability of the tools and resources used in the original project (Iqbal et al., 2016). Since the DSNS has not been tested in any other domain such as painting or music composition, we cannot say for certain these (or any of the previous) results would be consistent there as well. However, testing the original DSNS implementation (Iqbal et al., 2016) in another domain apart from chess (such as paintings and music) is a viable direction for further work in itself, not to mention also testing for the extended findings discovered here. Until neuroscientists know more about how creativity arises in the human brain, it will be difficult to explain why the DSNS approach works at all. Even then, it does not necessarily mean that creativity can only arise in the way it arises in humans. The answers to some of the open questions with regard to the DSNS may therefore require contributions from various disciplines such as neuroscience, neurology and even psychology. We are confident that in the future more progress will be made not only toward the 'how' questions (as in this article) but also the 'why' questions which are philosophically significant, if nothing else.

## 4      CONCLUSIONS

Computational creativity is an emerging sub-field of artificial intelligence. It allows computers to traverse the theoretical design space in ways that brute-force computing simply cannot. Previously thought to be limited only to the human brain, various techniques have been developed that demonstrate computers can indeed be creative by not only assessing but also producing objects of creative value. One of the most recent developments is the DSNS approach which has previously been shown to be able to compose chess problems of higher quality than random and state-of-the-art approaches. In this article, we have extended that work by showing that the output can be further improved by using higher quality materials. In particular, larger and clearer photographs in the seeding of the DSNS process. There are many other permutations that can and should be tested in order to obtain a clearer picture of what really works here and what does not but those are beyond the scope of the present work and will take time to build upon. Even so, objects generated using the DSNS as it stands now are no less commercially viable (Chesthetica, 2016) and any further improvements, however small, can still yield tangible benefits to practitioners and developers in the field.

# APPENDIX A: CHESTHETICA PROGRAM AND SETTINGS

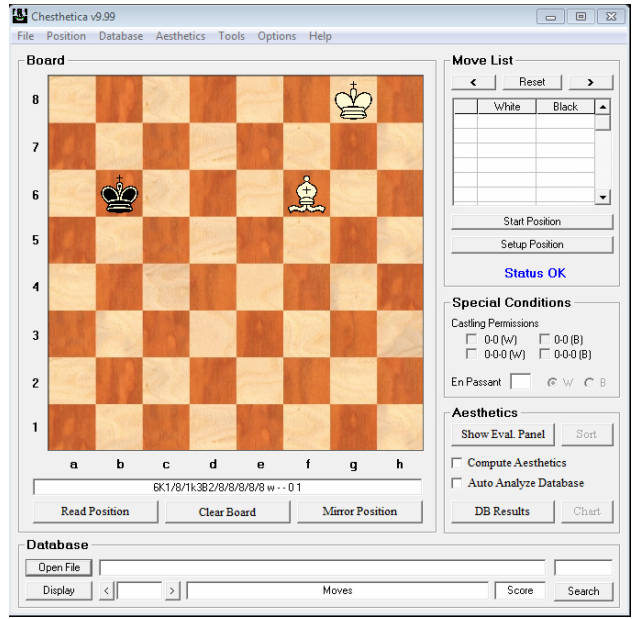

**Figure 6**: Main window of Chesthetica.

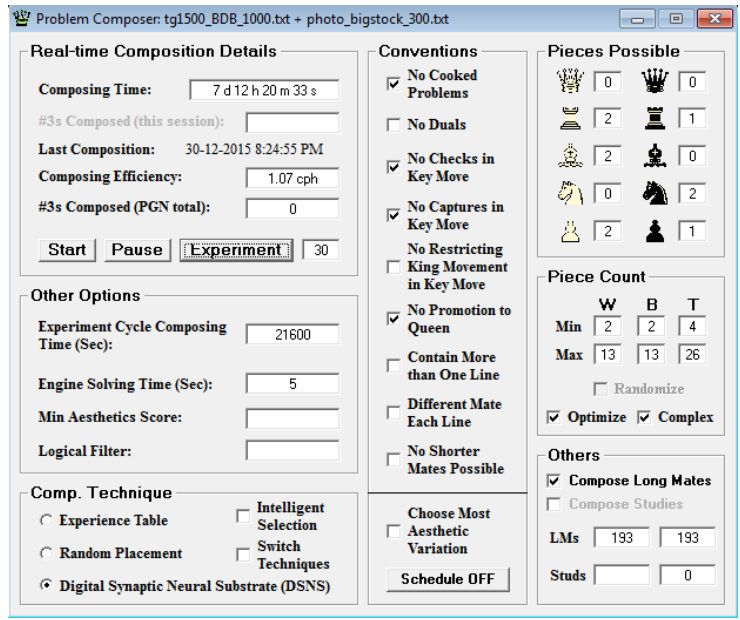

**Figure 7**: The automatic composer settings in Chesthetica used.